\documentclass[acmtog]{acmart} 
\pdfoutput=1
\renewcommand\footnotetextcopyrightpermission[1]{} 

\usepackage{colortbl}
\definecolor{Gray}{gray}{0.9}
\usepackage{lipsum}
\begin{document}

\title{Towards the Detection of Building Occupancy with Synthetic Environmental Data}


\author{Manuel Weber}
\email{manuel.weber@hm.edu}
\affiliation{%
  \institution{Munich University of Applied Sciences}
  \streetaddress{Lothstr. 64}
  \city{Munich}
  \state{Germany}
  \postcode{80335}
}

\author{Christoph Doblander}
\email{christoph.doblander@in.tum.de}
\affiliation{%
  \institution{Technical University of Munich}
  \city{Munich}
  \state{Germany}
}

\author{Peter Mandl}
\email{peter.mandl@hm.edu}
\affiliation{%
  \institution{Munich University of Applied Sciences}
  \streetaddress{Lothstr. 64}
  \city{Munich}
  \state{Germany}
  \postcode{80335}
}

\begin{abstract}
Information about room-level occupancy is crucial to many building-related tasks, such as building automation or energy performance simulation. Current occupancy detection literature focuses on data-driven methods, but is mostly based on small case studies with few rooms. The necessity to collect room-specific data for each room of interest impedes applicability of machine learning, especially data-intensive deep learning approaches, in practice. To derive accurate predictions from less data, we suggest knowledge transfer from synthetic data. In this paper, we conduct an experiment with data from a CO$_2$ sensor in an office room, and additional synthetic data obtained from a simulation.
Our contribution includes (a) a simulation method for CO$_2$ dynamics under randomized occupant behavior, (b) a proof of concept for knowledge transfer from simulated CO$_2$ data, and (c) an outline of future research implications. From our results, we can conclude that the transfer approach can effectively reduce the required amount of data for model training.
\end{abstract}


\maketitle
\thispagestyle{plain}
\pagestyle{plain} 

\section{Introduction}
Room-level occupancy in buildings provides crucial information to numerous use cases regarding, amongst others, building automation or building energy performance simulation. We differentiate between \textit{occupancy estimation}, i.e. the determination of an exact count of occupants in a room, and \textit{occupancy detection}, the binary discrimination between presence and absence. In both categories, a large number of models has been proposed over the last years. These include rule-based, stochastic and data-driven models. Recently, machine learning approaches have gained increasing interest. A recent study \cite{Carlucci.2020b} shows that 56\% of the models published between 2004 and 2019 are data-driven, 15\% use neural network techniques.
In addition to different model types, diverse sensing technologies can be found in literature. We classify sensing methods into \textit{intrusive} and \textit{non-intrusive}. Intrusive methods include additional sensors, such as light barriers or thermal cameras. Non-intrusive methods use existing data to predict occupancy states. \textit{Environmental sensing} of carbon dioxide rates, temperature etc. is considered non-intrusive, as climate sensors are already widely installed for the purpose of automated climate control.
Furthermore, climate sensors are more privacy-preserving than cameras, and, in contrast to transition-based measuring (e.g. with light barriers), they do not accumulate errors during the day.
Chen et al. (2017) propose a first deep learning model for building occupancy estimation based on environmental factors, outperforming several state-of-the-art models  \cite{Chen.2017b}. Deep learning allows the detection of occupancy at high precision without manual feature engineering. The challenge is that deep learning models require large quantities of room-specific labelled training data, which is not applicable in practice. 
Due to the high effort for collecting ground truth occupancy data for several days or weeks, models that provide good performance in a few test rooms cannot easily be applied to a variety of other rooms.
We propose to use \textit{transfer learning} to address this impediment. In our work, transfer learning is referred to in a transductive sense: Given equal learning tasks $T_s$ = $T_t$ for two domains (rooms) $R_s$ and $R_t$ with data drawn from different distributions within the same feature space, and given labeled data from the source domain $R_s$ and unlabeled or partially labeled data from the target domain $R_t$, transfer learning improves the training of $T_t$ using $R_s$ and $T_s$.
Some publications have already reported benefits of a knowledge transfer between different rooms.
Arief-Ang et al. (2017) extended a previously published seasonal decomposition model with a transfer approach, and used data recorded from a university office to effectively improve a model trained for a cinema hall \cite{AriefAng.2017}.
In a later approach \cite{Zhang.2019b}, transfer learning was successfully applied with a recurrent neural network (RNN),  which is favorable in terms of generalizability.
To the best of our knowledge, no deep transfer learning approach for environmental sensing-based detection of occupancy has been investigated to date. Moreover, synthetic data has not yet been considered for transfer, although physical simulation is common practice in the field of building technology. Tools such as EnergyPlus, TRNSYS and IDA~ICE provide simulations for various environmental variables with high accuracy \cite{Mazzeo.2019}. Regarding synthetic occupancy data, Chen et al. (2018) proposed an agent-based occupancy simulator \cite{Chen.2018c} to generate stochastic occupancy data in office buildings.
\begin{figure*}[!hbt]
  \includegraphics[scale=0.365]{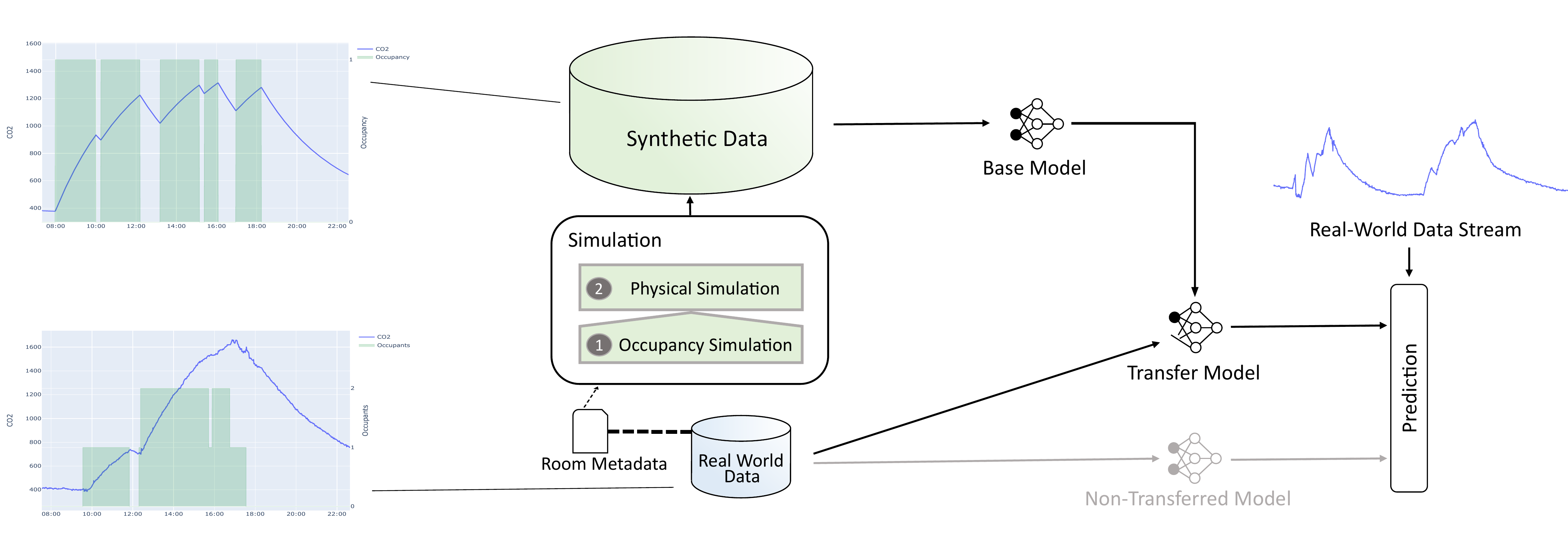}
  \caption{Simulation-aided approach to occupancy detection}
  \label{fig:approach}
\end{figure*}
\section{Approach}
We propose to leverage synthetic data generated from simulations to lower the need for costly real world data in context of occupancy detection. We conduct two simulations:
\begin{itemize}
 \item[(1.)] Occupancy Simulation: For occupant presence and possibly additional, relevant actions, such as window opening actions
 \item[(2.)] Physical Simulation: For environmental factors, such as CO$_2$
\end{itemize}
A simulation of occupants, and possibly of their actions (1.), provides the ground truth data that also serves as an input to a simulation of the physical environment (2.). In this work, we focus on CO$_2$ dynamics. Also other environmental factors such as temperature or humidity may be considered. A physical simulation requires metadata about the room, such as its volume and infiltration rate. Occupant simulations need to take into account the type of room, e.g. 2-person office or lecture room. Otherwise, occupant behavior may not be comparable. 
\pagebreak

\noindent When building a model for a concrete room of interest, we distinguish two approaches:

\begin{itemize}
 \item[(a)] run a preceding simulation based on metadata of the room of interest
 \item[(b)] use a general base model built from simulations on a variety of virtual rooms of the same room type
\end{itemize}
In this work, we use alternative (a) for a first proof of concept. It is illustrated in Fig.~\ref{fig:approach}. The two consecutive simulations (occupancy and CO$_2$) generate a large-scale synthetic dataset. This dataset is used to pre-train a base model that is able to fit the general behavior of CO$_2$ dynamics under human presence or absence. Room-specific behavior due to certain conditions such as sensor placement, infiltration, window size etc. is then learned in a transfer step. 
\section{Methods}
This section describes the simulation methods we applied to simulate occupancy as well as CO$_2$ rates. In Subsec.~\ref{section:occupancyDetection} we then introduce the model architecture used to detect occupancy.
\subsection{Occupancy Simulation}\label{section:occSim}
We simulate human occupancy in a naturally ventilated single office space of the size of a selected real world office. Occupancy behavior in offices can be subdivided into status transition events (e.g. arrival or final departure from work), random moving events (e.g. going to the bathroom) and meeting events \cite{Chen.2018c}.
As in \cite{Chen.2018c}, we combine the LIGHTSWITCH-2002 approach \cite{Reinhart.2004} for status transitions and Wang's (2011) Markov chain approach \cite{Wang.2011} for random moving.\\
\textbf{1. Status transition:}
For each day of simulation, we place an arrival at 8:00, a lunch break of 1h at 12:00, a departure at 18:00 and two 15-minute breaks at 10:00 respectively 15:00, with a random shift of ${\pm}$15min for each event \cite{Reinhart.2004}. The timespans determine basic occupancy. Within these timespans, random moving events are simulated.\newline
\textbf{2. Random moving:}
Throughout all discrete time steps at basic occupancy, we use a Markov chain to successively determine the state of occupancy $occ[t]\in\{0, 1\}$ at time t, depending on the previous state $occ[t-1]$ and a transition probability \cite{Wang.2011}. Transition probabilities are picked from the transition matrix $P$ depicted in Eq.~\ref{eq:transitionProbabilities}. To consider a variety of different occupant types, we update $P$ for each simulated day. The used sojourn times, $s_0$ for transitions from absence (0) to presence (1), and $s_1$ vice versa, are randomly selected within the following bounds.
\vspace{0.2mm}
\begin{align}
  \label{eq:transitionProbabilities}
  \left. \begin{aligned}
    P = \begin{bmatrix}
    1-(\frac{1}{s_0}) & \frac{1}{s_0}\\
    \frac{1}{s_1} & 1-(\frac{1}{s_1})
    \end{bmatrix}
  \end{aligned} 
  \right. \begin{aligned}
  \ &&~10min~\leq~s_0~\leq~60min~\\
  \ &&30min~\leq~s_1~\leq~180min
  \end{aligned}
\end{align}

\noindent\textbf{3. Window opening behavior:}
We decided to use a similar Markov chain to determine the window state w[t]$\in$\{0, 1\} where 0 denotes that all windows or openings are closed, and 1 represents a state of ventilation. We set the time bounds to reflect different ventilation behavior to 
$60min\leq s_0^w\leq8h$ and $5min \leq s_1^w \leq 30min$. Windows can be open also in case of absence due to random moving.

\subsection{Carbon Dioxide Simulation}\label{section:co2Sim}
The change in CO$_2$ rate per time step within a room of volume V is calculated using Eq.~\ref{eq:massBalance}. The formula is adopted from \cite{Parsons.2014}.
\vspace{0.2mm}
\begin{equation}
  \label{eq:massBalance}
  \frac{dc(t)}{dt}=\frac{\dot{m}(t)}{V}
  (c_{out} - c(t))+\frac{G(t)}{V}
\end{equation}
\noindent
$c(t)$ is the indoor CO$_2$ concentration at time t, c$_{out}$ is the outdoor CO$_2$ concentration,
$\dot{m}(t)$ is the current mass flow rate, and $G(t)$ is the amount of CO$_2$ generated by human occupants present at the moment. Applying Eq.~\ref{eq:massBalance}, 
we successively calculate CO$_2$ level values for time steps of one second, which can be later aggregated to the adequate granularity. For simplification purposes, air pressure and outdoor CO$_2$ are considered constant.
The infiltration rate $\dot{m}(t)$ consists of a steady infiltration $\dot{m}_{inf}(t)$ independent from human occupancy, and a ventilation rate $\dot{m}_{vent}(t)$ determined by occupants' actions, especially window opening events: $\dot{m}(t) = \dot{m}_{inf}(t) + \dot{m}_{vent}(t)$.
To simplify, we use a constant value $\dot{m}_{inf}(t)$ = $\dot{m}_{inf}$.
Also, we consider a detailed calculation of $\dot{m}_{vent}(t)$ under consideration of varying window counts, -types and -sizes, as well as opening angles and air pressure differences, as unnecessary complex in a first proof of concept. Hence, whenever the window state is 1, we use a multiple of $\dot{m}_{inf}$ instead: $\dot{m}(t) = \dot{m}_{inf} \cdot vm$. A multiplier $vm$ is randomly chosen from the range [10, 100] at each window opening event, and set to 1 if windows are closed.
$\dot{m}_{inf}$ is calculated from the volumetric air flow rate $\dot{V}_{inf}$ by multiplication with the mass of air: $\dot{m}_{inf}$ = $\dot{V}_{inf} \cdot m_{air}$. The value of m$_{air}$ is set to 1.2754 g/l according to the IUPAC standard for the mass of dry air at standard temperature and pressure.
$\dot{V}_{inf}$ can be estimated by using CO$_2$ as a tracer gas, fitting a theoretical decay curve to the actual decay in CO$_2$ during a period of non-occupancy \cite{Parsons.2014}.
G(t), defined in Eq.~\ref{eq:humanCO2Generation}, is
the product of the number of present occupants n(t) (in our case 0 or 1), an average CO$_2$ generation rate of a single occupant $g_{occ}$, 
the mass of CO$_2$ $m_{co2}$
and a unit translation term. For simulating intervals of one second, the term is divided by 60.
\begin{equation}
    \label{eq:humanCO2Generation}
    G(t) = n(t) \cdot g_{occ} \cdot m_{co2} \cdot \frac{1000}{60},
    \hspace{2mm} n(t) \in \{0, 1\}
\end{equation}
We choose $g_{occ}$ = 0.24 l/min, which is the mean value measured by \cite{Reiff.2012} for a person of an age between 21 an 28 standing at a desk. 0.18, the other value from \cite{Reiff.2012}, achieves lower results in our experiment.
\vspace{-2mm}
\subsection{Occupancy Detection}
\label{section:occupancyDetection}
\begin{figure}[!hbt]
  \includegraphics[width=\linewidth]{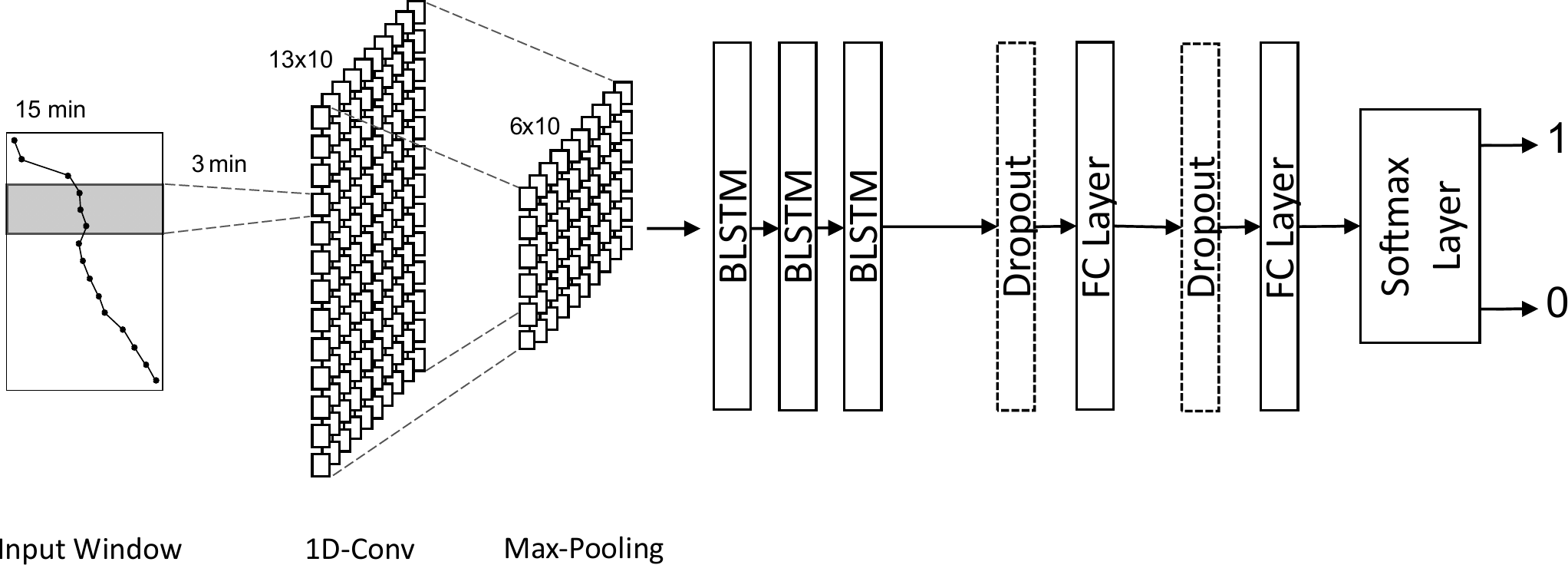}
  \caption{Deep learning model architecture based on \cite{Chen.2017b}}
  \label{fig:modelArchitectur}
\end{figure}

To detect occupancy, we apply the model architecture from \cite{Chen.2017b}, see Fig.~\ref{fig:modelArchitectur}. It combines a convolutional network with a deep bidirectional long short-term memory (DBLSTM). 
We downsample the data to a 1-min-granularity aggregating by the mean, and then use a sliding 15min-window on the input data stream. A one-dimensional convolutional layer (1D-Conv) and a max-pooling layer perform an automated feature extraction. As in \cite{Chen.2017b}, we use a filter size of 3, and a pooling factor of 2. We reduce the number of filters to 10, as we use CO$_2$ as a single input factor. The number of neurons in the following bidirectional long short-term memory (BLSTM) layers (200, 150 and 100 neurons) and the fully connected (FC) layers (300 and 200 neurons) are selected as proposed in \cite{Chen.2017b}, as well as the masking probabilities (0.5 and 0.3) for the dropout applied before the first and second FC layer for regulation purposes. A final softmax activation layer discriminates between presence (1) and abscence (0).

\section{Experimental Setup}
\subsection{Datasets}
\textbf{Real world dataset:}
For a preliminary study, we measured the CO$_2$ rate in a 2-person office of a German university building located in Munich throughout 7 complete working days. A \textit{Sensirion~SCD30} non-dispersive infrared (NDIR) sensor was placed in the center of the room, and a measured CO$_2$ value was stored every 5 seconds. According to the datasheet, the sensor provides a resolution of 1~ppm with an accuracy of ${\pm}$30 ppm +3$\%$. The CO$_2$ data was collected between 10 and 19 March 2020. Occupancy ground truth data was manually recorded by the occupants over the observation period.\linebreak
\textbf{Simulated dataset:}
In addition to this, we conducted a CO$_2$ time series simulation according to the methods previously described in Sec.~\ref{section:occSim} and \ref{section:co2Sim} for a total of 500 artificial working days in a room with the same volume as the real world office ($V$ = 77.5 m³). A natural infiltration rate and outdoor CO$_2$ concentration were estimated fitting Eq.~\ref{eq:massBalance}
to observed CO$_2$ values from one unoccupied night not used for the dataset by minimizing the mean squared error (MSE). Consequently, they were set to realistic values of $\dot{V}_{inf}$ = 0.0046 m³/s and c$_{out}$ = 360 ppm. Table \ref{tab:datasetOverview} summarizes the properties of the two datasets.
Note that two occupants share the office, and guests may visit. Our simulation is only intended to provide basic information for a single-person scenario.
\begin{table}[!hbt]
\caption{Dataset Overview}
  \label{tab:datasetOverview}
  \begin{tabular}{ccc}
    \toprule
    & Real World Dataset & Simulated Dataset\\
    \midrule
    Dataset Size & 7 working days & 500 working days \\
Time Granularity & 5 sec & 1 sec \\
    Occupancy Values & [0, 3], mostly [0, 2] & [0, 1] \\
    Presence Rate & 29.28$\%$ & 25.73$\%$\\
    CO$_2$ Value Range & [338, 1749] ppm & [360, 1483] ppm \\
    \bottomrule
  \end{tabular}
\end{table}

\subsection{Experiments}
We conducted a first study to show the positive effects of transfer learning from synthetic CO$_2$ data. First, 400 days of our simulated dataset were used to train a synthetic base model. The remaining 100 days were used to report the performance of the base model on simulated data.
Using only small amounts of real world training data, we then trained two models: (1) a conventional deep learning model as described in Sec.~\ref{section:occupancyDetection}, and (2) a transfer model with equal architecture and hyperparameters, using the same real world data, but based on the synthetic base model.
The conventional model was trained using a uniform random initializer for model weights.
\begin{table*}[ht]
  \caption{Comparison of transfer model, non-transferred model and LR baseline for different training data extents*}
  \label{tab:Results}
  \begin{tabular}{llccccc}
    \toprule
    & Training Data & 1 Day & 2 Days & 3 Days & 4 Days \\
    \midrule
    Transfer Model 
    & Accuracy
        &  \textbf{0.875} \scriptsize{(${\pm}$\textbf{0.057})}
        &  \textbf{0.915} \scriptsize{(${\pm}$\textbf{0.020})}
        &  \textbf{0.929} \scriptsize{(${\pm}$\textbf{0.014})}
        &  \textbf{0.933} \scriptsize{(${\pm}$\textbf{0.015})}\\
    & F1 Score
        &  0.764 \scriptsize{(${\pm}$0.171)}
        &  \textbf{0.863} \scriptsize{(${\pm}$\textbf{0.035})}
        &  \textbf{0.888} \scriptsize{(${\pm}$\textbf{0.027})}
        &  \textbf{0.898} \scriptsize{(${\pm}$\textbf{0.028})}\\
    & Epochs
        &  27.2 \scriptsize{(${\pm}$16.6)}
        &  26.1 \scriptsize{(${\pm}$14.4)}
        &  22.6 \scriptsize{(${\pm}$14.1)}
        &  19.9 \scriptsize{(${\pm}$11.9)}\\
    \cmidrule{2-6}
    Non-Transferred Model 
    & Accuracy
        &  0.715 \scriptsize{(${\pm}$0.086)}
        &  0.874 \scriptsize{(${\pm}$0.040)}
        &  0.895 \scriptsize{(${\pm}$0.035)}
        &  0.914 \scriptsize{(${\pm}$0.029)}\\
    & F1 Score
        &  0.565 \scriptsize{(${\pm}$0.162)}
        &  0.791 \scriptsize{(${\pm}$0.065)}
        &  0.821 \scriptsize{(${\pm}$0.071)}
        &  0.859 \scriptsize{(${\pm}$0.053)}\\
    & Epochs
        &  91.3 \scriptsize{(${\pm}$54)}
        &  123 \scriptsize{(${\pm}$50.2)}
        &  100.8 \scriptsize{(${\pm}$46)}
        &  105.2 \scriptsize{(${\pm}$34.4)}\\
    \cmidrule{2-6}
    Logistic Regression (LR) & Accuracy 
        & 0.860 \scriptsize{(${\pm}$0.073)}
        & 0.879 \scriptsize{(${\pm}$0.058)}
        & 0.900 \scriptsize{(${\pm}$0.035)}
        & 0.920 \scriptsize{(${\pm}$0.016)}\\
    & F1 Score
        & \textbf{0.767} \scriptsize{(${\pm}$\textbf{0.112})} 
        & 0.803 \scriptsize{(${\pm}$0.080)} 
        & 0.829 \scriptsize{(${\pm}$0.056)} 
        & 0.866 \scriptsize{(${\pm}$0.038)}\\  
    \bottomrule
  \end{tabular}\\
  \vspace{2mm}
  \small{*mean values (${\pm}$ standard deviations), best scores are highlighted in bold}
\end{table*}
Transfer learning was carried out by using the model weights of the base model for initialization instead, and retraining all layers.
We evaluated with amounts of 1, 2, 3 and 4 days of training data. A cross validation was applied, using each of the 7 days in our dataset (and accordingly each 2, 3 or 4 consecutive days) for training in one iteration, and the remaining days as test data. Each iteration was repeated 10 times with a new seed value for randomization in initial weight generation and shuffling of input sequences.
All experiments were carried out on an \textit{Nvidia GeForce Tesla V100 SXM2} GPU, using the TensorFlow framework with Keras.
For optimizer, learning rate, batch size and validation split, we chose RMSprop, 0.001, 70 and 0.2. Early stopping was applied after 20 epochs of no improvement in validation loss.
To compare the results, we calculate the detection accuracy, which is defined as the number of correctly predicted occupancy states divided by the total number of predictions. As a second metric, we use the F1 score.
\section{Results}
Table \ref{tab:Results} reports the mean and standard deviation values of accuracies, F1 scores and training epochs under different extents of training data, in both the transfer and non-transfer setting. Additionally, results are compared to a logistic regression (LR) baseline. The pre-trained transfer model clearly outperformed the non-transferred model, with a higher impact the less training data was used. Vice versa, only half of the training data was required to achieve a similar accuracy as in the non-transfer setting: 0.87 was reached with one instead of two days, and 0.91 with two instead of four days of training data. Besides a substantial improve in accuracy and F1 score, also the standard deviation was reduced in most cases. Hence, the transfer approach can also improve model robustness.
As data was scarce, the deep learning model was unable to reach its full potential, and performed even slightly worse than an LR classifier. The transfer model, in contrast, showed superior results in nearly all cases.
Regarding training times, the transfer model was able to be trained in only a fraction of the training epochs. The number of epochs reported in Table \ref{tab:Results} indicate after how many training epochs validation loss reached a minimum with respect to all previous and 20 subsequent epochs. By average, 91 epochs were trained for the non-transferred model with one day of training data, and above 100 with multiple days of training data. In contrast to this, transfer learning allowed a validation loss convergence after less than 30 epochs. However, this reduction of training times results from the additional upfront effort for pre-training. The effect is only beneficial if the base model is reused for multiple rooms. Pre-training required 30 epochs on data from 400 synthetic training days. The resulting base model achieved an accuracy of 0.981 and an F1 score of 0.963 on 100 simulated days. Without the subsequent transfer step, the base model by itself was not able to make accurate predictions on the real world test dataset. Simulations do not replace but reduce the need for data collection.

\section{Conclusion \& Future Research}
In our experiment, we have demonstrated that transfer from synthetic data can effectively improve model performance and robustness regarding occupancy detection. The results encourage the use of simulations in this field of limited real world data to enable deep learning in practice.
We see future research in finding a concrete method considering how to generate adequate synthetic data, and how to accomplish transfer.
It should also be investigated whether a large, generalized base model, as it is common in the field of image processing for instance, may be beneficial for a variety of different rooms. 
We aim to prepare a broader base model and evaluate on multiple room types.
Furthermore, we are planning to investigate other model architectures and their ability to transfer knowledge. For the purpose of demonstration, in this work several constant values and arbitrary boundaries were used in the simulation. We want to overcome these limitations, and also consider predicting the number of present occupants.

\begin{acks}
\begin{anonsuppress}
Special thanks go to the \textit{OpenPower@TUM} project and the industry partner IBM for providing us with the computing infrastructure for our experiments.
\end{anonsuppress}
\end{acks}

\bibliographystyle{ACM-Reference-Format}
\bibliography{bibliography}


\end{document}